\documentclass[11pt]{article}
\usepackage{coling2020}
\usepackage{times}
\usepackage{url}
\usepackage{latexsym}
\usepackage{graphicx} 

\usepackage{amsmath}
\usepackage{amssymb}
\DeclareUnicodeCharacter{2212}{-}
\DeclareUnicodeCharacter{00D7}{\times}

\usepackage{caption}
\usepackage{subcaption}

\usepackage{multirow} 
\usepackage{booktabs} 
\colingfinalcopy 
\usepackage{dcolumn}

\usepackage{wrapfig} 

\title{E.T.: Entity-Transformers \\ Coreference augmented Neural Language Model for richer mention representations via Entity-Transformer blocks}

\author{Nikolaos Stylianou \\
  Aristotle University of Thessaloniki \\
  School of Informatics \\
  Greece \\
  \texttt{nstylia@csd.auth.gr}\\ \And
  Ioannis Vlahavas \\
  Aristotle University of Thessaloniki \\
  School of Informatics \\
  Greece \\
  \texttt{vlahavas@csd.auth.gr} \\}

\date{}

\begin{document}
\maketitle
\begin{abstract}
    In the last decade, the field of Neural Language Modelling has witnessed enormous changes, with the development of novel models through the use of Transformer architectures. However, even these models struggle to model long sequences due to memory constraints and increasing computational complexity. Coreference annotations over the training data can provide context far beyond the modelling limitations of such language models. In this paper we present an extension over the Transformer-block architecture used in neural language models, specifically in GPT2, in order to incorporate entity annotations during training. Our model, GPT2E, extends the Transformer layers architecture of GPT2 to Entity-Transformers, an architecture designed to handle coreference information when present. To that end, we achieve richer representations for entity mentions, with insignificant training cost. We show the comparative model performance between GPT2 and GPT2E in terms of Perplexity on the CoNLL 2012 and LAMBADA datasets as well as the key differences in the entity representations and their effects in downstream tasks such as Named Entity Recognition. Furthermore, our approach can be adopted by the majority of Transformer-based language models.
\end{abstract}

\section{Introduction}

Language modelling is the task of transforming individual words into vector representations based on the context they appear in. Hence, distant term dependencies are an inherited issue within the task. Language models always seek for smart approaches towards incorporating context from longer distances as it allows for better representations compared to their limited context counterparts. Intuitively, imagine attempting to start reading a novel series from the second book onward, with no information about the first. The amount of information previously missed is something that cannot be acquired. However, this is the case with most language models. While an understanding of the words is present due to the contextual information at each word's occurrence, entity information that are in distant text are lost or not transferred. 

Until recently, Recurrent Neural Networks (RNNs), and specifically  Long Short-Term Memory (LSTM) networks, have been the core of all the state-of-the-art approaches \cite{mccann2017learned-cove,peters2018deep-elmo}. Thanks to the Transformers architecture \cite{vaswani2017attention}, through the use of attention mechanisms, models such as XLNet \cite{yang2019xlnet}, GPT \cite{radford2019language} and BERT \cite{devlin2018bert} can account for even longer sequences. However, the computational limitations of the multi-head attention in the architecture make it hard to increase the contextual information in such models \cite{tay2020efficient}. As a result, research has been focused on introducing variations to the transformer architecture, with focus on the multi-head attention mechanism, in order to alleviate part of the computational cost and increase the contextual information available to models. 

In this paper we present a novel approach, that makes use of coreference information during training a language model via our Entity-Transformer architecture, which extends the original Transformer block in Transformer-Based language models. To that end, we incorporate the important entity information that would otherwise be unreachable for the model. As a result, we effectively boost the representations of the entity mentions, where entity information is present, without hindering the performance of the language model where entities are not present. 

In our experiments, we extend the GPT2 architecture to formulate our model, named GPT2E and train it on the CoNLL-2012 dataset \cite{pradhan2012conll} using the annotated coreference information. We evaluate the model's performance in terms of Perplexity on the ConLL 2012 and the LAMBADA \cite{paperno2016lambada} datasets and showcase the effects of such training on the word representations as well as on the downstream task of Named Entity Recognition (NER) using the CoNLL 2012 dataset. To that end, we compare GPT2E's performance to a base model (GPT2) when trained on the same data, to highlight the effects of coreference information when paird with our Entity-Transformer architecture. 

\section{Related Work}
\label{sec:related-work}
In the last decade, the field of Neural Language Modelling has witnessed enormous changes. 
With pretrained neural language models being the current go-to approach in all NLP reserach, a variety of methods models have been developed. We distinguish two major categories:

\paragraph{General purpose language models.}
Steady improvements have been achieved to this field with the use of deep RNNs and pre-training on a large number of training data \cite{mccann2017learned-cove,peters2018deep-elmo}. With Transformers, language models have been able to capture longer linguistic structures without the use of RNNs and surpass their RNN counterparts by a big margin \cite{radford2018improving,devlin2018bert}. Recent research has focused on ways of taking advantage of more context \cite{yang2019xlnet,fan2020accessing-feedback-transformers} and introducing effective methodologies to scale up the models and train them \cite{radford2019language,shoeybi2019megatron,rosset2019turing,brown2020language}. 

\paragraph{Language modelling with entity decisions.}
YangLM \cite{yang2016reference} was the first to incorporate entity decisions to a language model by introducing learnable entity embeddings. Alternative entity handling mechanisms are introduced in both EntityNLM \cite{ji2017dynamic} and SetLM \cite{kunz2019entity} in addition to a length variable for EntityNLM. All of the aforementioned approaches are RNN-based and hence their performance is expected to be sub-par to Transformer based models. Furthermore, \cite{kunz2019entity} concludes that language models handling entity decisions do not improve in performance with the addition of more hidden units and that the source data is of limited number and of specific genre which do not highlight the benefits of explicit entity information. \newcite{clark-etal-2019-bert}, through attention head probing, experimentally proves that BERT does model anaphoric phenomenon in the form of antecedent selection, with attention heads directly attending to the respective mention's antecedent. However, these information are not explicitly used to further enhance the model. Furthermore, ERNIE \cite{zhang-etal-2019-ernie}, which uses knowledge graphs to infuse entity information to the model, only does so for named entities, completely ignoring pronouns and nominal mentions.

\section{Our approach}
In order to incorporate coreference information to a language model, we require training and testing data with entity information present and a mechanism to handle existing and non-existing entities. To that end, our proposed model, GPT2E, is based on the GPT2 language model, with changes to the Transformer block and an entity handling mechanism, which are described in the following subsections. As a result, GPT2E is a combination of multi-layer Entity-Transformer decoder blocks. The model applies multi-headed self-attention operations over the input tokens, position-wise feed-forward transformations, and entity-based attention operations. The model architecture can be described as follows: 
\begin{equation}
\begin{split}
     h_{0} = UW_{e} + W_{p} \\
     h_{l} = \texttt{entity\_transformer\_block}(h_{l-1},E) \forall_{i} \in [1,n]  \\
     P(u) = \texttt{softmax}(h_{n}W_{e}^{T})
\end{split}
\end{equation}

where $U = (u\__{k}, \dots, u\__{1})$ is the context vector of tokens, $n$ is the number of layers, $W_{e}$ is the token embedding matrix, $W_p$ is the position embedding matrix and $E$ is the context vector of entity representations. 

\subsection{Entity-Transformer block}

Entity-Transformer (ET) blocks are extensions of the transformer blocks used in GPT2, designed to handle entities in the form of vectors of shape $E_{i} \in \mathbb{R}^{1 \times d_{embd}}$, where $d_{embd}$ is the embedding dimension the model outputs. Effectively, the entity representations are used directly inside the ET blocks. 

The input representation first goes through a layer normalization \cite{ba2016layer} and a masked multi-head self attention layer \cite{vaswani2017attention}, followed by a residual connection \cite{he2016identity}. The output of the residual connection is then used in a layer normalization and position-wise feed foward layer followed by another residual connection. The final residual output is used in the entity attention layer before it is forwarded outside of the Entity-Transformer block.

\begin{wrapfigure}{r}{0.6\linewidth}
\centering
\includegraphics[scale=0.4]{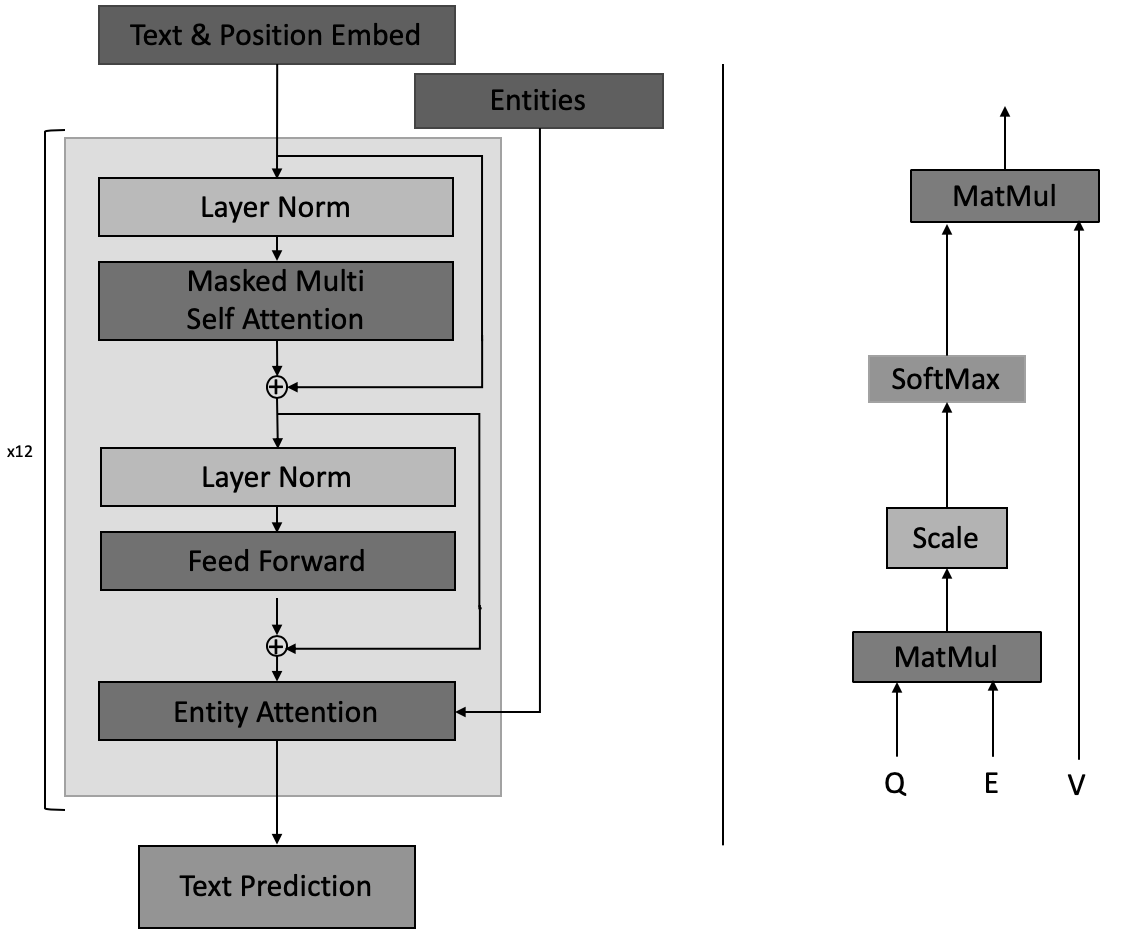}
\caption{\textbf{(left)} Entity-Transformer Block \textbf{(right)} Entity Attention mechanism}
\label{fig:et_block}
\end{wrapfigure}

The entity attention layer is an adaptation of the masked multi-head self attention layer which considers Entities (E) as the Key (K) value in the Query (Q), Key (K), Value (V) attention mechanism scheme. The architecture of the Entity-Transformer blocks and the entity attention mechanism used are shown in Figure \ref{fig:et_block}.

\subsection{Entity handling mechanism}
We maintain a persistent set of entities $\mathcal{E}$, that holds the hidden representation of the last entity's mention from our model. Each entity representation $E_{i}$ is initialised as a vector of ones, which allows for minimal noise in the first occurrence of the entity. Tokens that are not part of the entity mention have a consistent entity representation $E_{\emptyset}$, as a vector of ones, similar to unseen entity mentions.

During each training step, $E_{i}$ takes the latest value of the respective entity's latest hidden representation from $\mathcal{E}$ and is updated to the new value at the end of each step. These entity representations are handled with the use of Entity-Transformer blocks. 
The final hidden representation of the input token, after it is affected by the previous entity representation $E_{i}$, is considered to be the new entity representations and replaces $E_{i}$ in $\mathcal{E}$.  

\section{Experiments}
Our approach is evaluated in two steps. 
First we evaluate our GPT2E language model, in comparison with a GPT2 model, trained on CoNLL 2012 and evaluated on both CoNLL 2012 and LAMBADA datasets. We then use the trained models to extract word representations for entity mentions based on the coreference annotations in text and measure the differences of such representations. For NER, we use the language models to extract word representatios and train the same baseline model on the CoNLL 2012 dataset.

\subsection{Setup}
In our experiments we use the GPT2-small configuration with 117M parameters, 12 heads and 12 layers for both GPT2 and GPT2E.
Both models use a Byte-Pair Encoder to process the input, a learning rate of 2e-5 and train for 10e5 steps, with validation every 10e3 steps. We use a batch size of 1, to highlight the effect of entity updates in the system, as the entity representations are only updated at the end of each training step.

After training, we compute the differences between the representations of all entity mentions in the coreference clusters as derived from GPT2 and GPT2E. Consequently, we conduct experiments with no contextual information for each word and we also distinguish the results between using and not using entity information. We perform these experiments separately for all entities in the dataset and present the average score for different type of words based on their part-of-speech tags. 
 
The NER models are based on the \newcite{lample2016neural} architecture. However, our models use only word embeddings from the pre-trained GPT2 and GPT2E models respectively, removing the character embeddings to eliminate any information input apart from the coreference-trained representations. We use a hidden size of 512 for the Bidirectional LSTMs, 0.5 dropout \cite{srivastava2014dropout} between layers and a learning rate of 0.0001 with 0.9 decay per epoch with Adam \cite{kingma2014adam}. We trained our models for 20 epoches, with early stopping and a batch size of 32. 

All the experiments were run on a computer with a single Titan V 12GB graphics card, 32GB of memory and an Intel i7-8700 processor.

\subsection{Datasets and Preprocessing}
\label{sec:datasets_preprocessing}
We chose the English CoNLL-2012 dataset for training, which is based on the OntoNotes 5.0 corpus \cite{weischedel2011ontonotes} and contains over 1.3 million words with 35,143 entity mentions in the training set and 170 thousand words with 4,532 entity mentions in the test set making it the most suitable dataset for training a language model with coreference annotations. In the dataset common nouns, pronouns and proper nouns contribute 90\% of the words in both train and test English sets. For our out of domain evaluation we chose the LAMBADA dataset. This choice was based on the premise that the dataset is primarly used for word predictions requiring broad discourse context and that the target words are mostly proper nouns and common nouns (85\% fo the total target words). As a result, we expect that the importance of an entity-centric language model would be better displayed in such a scenario. 

As we utilize the CoNLL-2012 dataset for both the Language Modelling task and the NER task, we formulate the data in two different ways.
\begin{table}[!ht]
    \centering
    \caption{Data example from the CoNLL 2012 dataset, as formated for the task.}
    \begin{tabular}{c|ccccccccccccccccccccc}
        \hline
        $X_{1:11}$ &`` & The & U.S. & underestimated & Noriega & all & along & '' & says & Ambler & Moss \\
        $E_{1:11}$ &  $\emptyset$ & 73 & 73 &  $\emptyset$ & 82 &  $\emptyset$ &  $\emptyset$ &  $\emptyset$ &  $\emptyset$ & 50 & 50 \\
        \hline
    \end{tabular}
    \begin{tabular}{c|ccccccccccccccccccccc}
         $X_{12:23}$ & a & former & Ambassador & to & Panama & . & `` & He & has  & mastered & the & art \\ 
         $E_{12:23}$ & 50 & 50 & 50 & 50 & 50 &  $\emptyset$ &  $\emptyset$ & 82 &  $\emptyset$ &  $\emptyset$ &  $\emptyset$ &  $\emptyset$ & \\
    \hline
    \end{tabular}
    
    \label{tab:data_example}
\end{table}

For Language Modelling, we formulate our data in a similar manner with \newcite{ji2017dynamic}, as seen in Table \ref{tab:data_example}. Specifically, for each token we also introduce a second variable ``$E$'' which indicates the entity in which the token is part of, using the gold coreference annotations, with a special ``$\emptyset$'' for tokens that are not part of an entity. For the CoNLL dataset, we populate $E$ with the golden entities from the coreference resolution shared task. For the LAMBADA dataset we use the $\emptyset$ for all tokens. In comparison to the original data formulation described in \newcite{ji2017dynamic}, we opted to not use the $L$ variable to denote the entity length (i.e. the number of remaining tokens in the entity mention) as it's main use is enable entity mention prediction, which we do not attempt at this stage. We use Byte Pair Encoding (BPE) \cite{sennrich-etal-2016-neural} for the final input representation of the word instances, similar to GPT2. 

For NER, we formulate the data in a IOB format to facilitate a similar model architecture as described in \newcite{lample2016neural}, using the gold named entities of the dataset, including nested entities.
\section{Results}

To evaluate the results of our Entity-Transformers architecture and the effects of corereference annotations to language modelling, we measure the change in performance of the language model using Perplexity (PPL). Furthermore, we compute the average difference of the representations between mentions of the same entity of the GPT2E model, between each entity mention between GPT2 and GPT2E and between non-entity mentions of the same words using cosine similarity. Furthermore, we use micro-average Precision, Recall and F1 scores for the evaluation of our NER models.

\begin{figure}[ht]
    \centering
    \includegraphics[width=.9\textwidth]{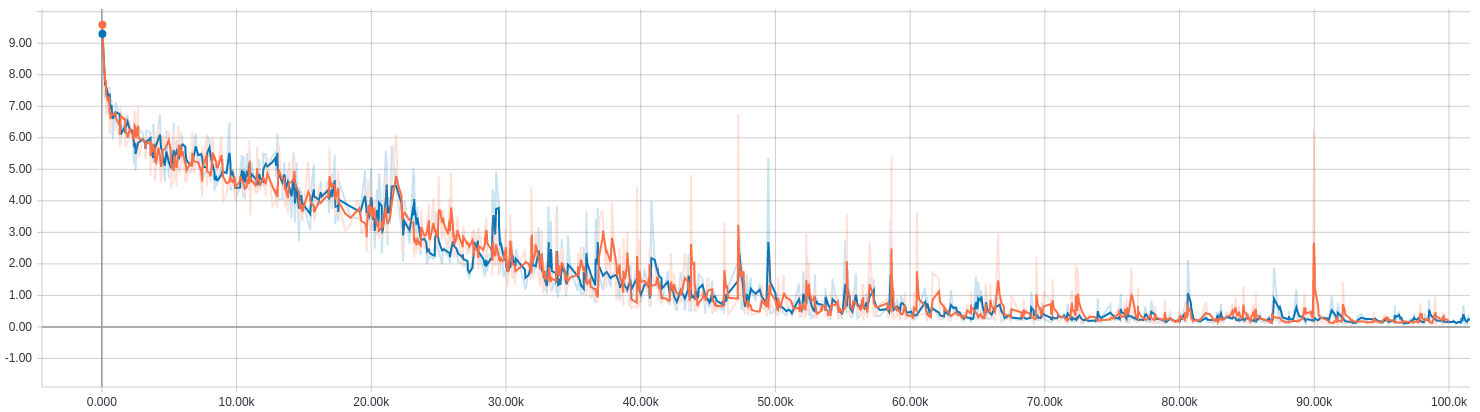}
    \caption{Training loss per step on the CoNLL 2012 dataset.}
    \label{fig:train_loss}
\end{figure}

\begin{figure}[ht]
    \centering
    \includegraphics[width=.9\textwidth]{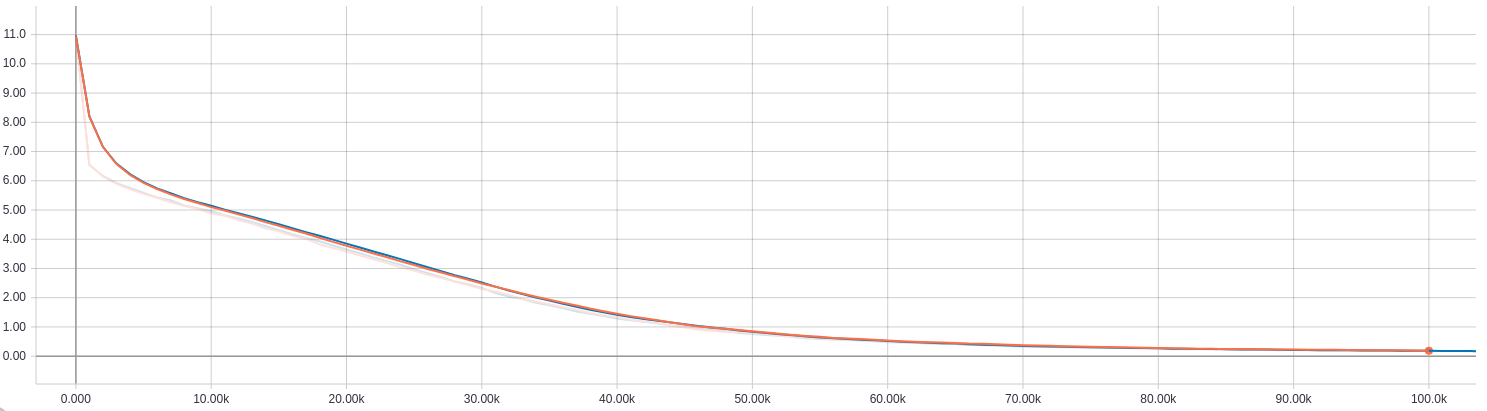}
    \caption{Validation loss per step on the CoNLL 2012 dataset.}
    \label{fig:val_loss}
\end{figure}

For Language modelling, Table \ref{tab:conll_ppl}, shows the training and validation losses of GPT2 and GPT2E, as well as the Perplexity of the models after 10e5 training steps. The gradual changes in training and validation losses, measured every 10e3 steps, are illustrated in Figures \ref{fig:train_loss} \& \ref{fig:val_loss} with GPT2 model in orange and GPT2E model in blue colours respectively. Similarly, Table \ref{tab:lambada_acc} highlights the performance difference between the two trained models on the LAMBADA dataset. As both models are trained on a very limited dataset compared to other language models, we are not comparing performance in terms of accuracy.

\begin{table}[!ht]
    
    \begin{minipage}{.65\linewidth}
    
        \caption{Perplexity and Validation loss \\on the CoNLL 2012 dataset}
        \begin{tabular}{|c|c|c|c|c|c|l|}
\hline
\multirow{2}{*}{Process} & \multicolumn{3}{c|}{GPT2E} & \multicolumn{3}{c|}{GPT2} \\ \cline{2-7} 
                         & PPL    & Loss    & \begin{tabular}[c]{@{}l@{}}Time\\ per step\end{tabular}    & PPL    & Loss    & \begin{tabular}[c]{@{}l@{}}Time\\ per step\end{tabular}   \\ \hline
Training & 5.52 & 1.71 & 0.290s & 4.80 & 1.57 &  0.298s  \\ \hline
Validation & 1.20  & 0.187 &  0.290s & 1.19 &  0.184 &  0.298s   \\ \hline
\end{tabular}
    \label{tab:conll_ppl}
    \end{minipage}%
    \begin{minipage}{.4\linewidth}
        \caption{Perplexity performance \\on the LAMBADA dataset}
        \begin{tabular}[!htb]{|c|c|c|}
            \hline
             Model & Perplexity   \\ \hline
             GPT2E &  196.81  \\ \hline
             GPT2 &  219.97  \\ \hline
        \end{tabular}
        \label{tab:lambada_acc}
    \end{minipage} 
\end{table}

In terms of Perplexity, the models show similar performances on the CoNLL 2012 dataset, while having a slight advantage at the LAMBADA dataset. The slight improvement in Perplexity of the GPT2E model over the GPT2 on the LAMBADA dataset is attributed to the target words' part-of-speech type. As described in Section \ref{sec:datasets_preprocessing}, the target words of the LAMBADA dataset are mostly proper nouns and common nouns and the majority of the training mentions in the CoNLL-2012 dataset are of the same type. This behaviour is consistent with the expectations of the performance of an entity-centric language model. Both GPT2 and GPT2E models show a remarkably low Perplexity compared to EntityNLM, YangLM and SetLM of reported Perplexity 161.64, 114 and 107 respectively. However, these language models are RNN based, and gap between them is attributed to the Transformers architecture and the relatively small size of the CoNLL-2012 dataset.
The added complexity of calculating the entity representations and using the Entity-Transformer blocks is contributing to 0.008 seconds per step in both training and evaluation, adding up to an additional 12 min and 6 seconds, a 2\% increase in time for the complete training process.

To compare the changes in the entity mention representations when using coreference information during training we conducted a series of experiments, taking into account the existence or absence of coreference annotation. Specifically, for both models, for each entity we calculate the average similarity of its mentions with the other entity mentions, with and without the use of entity representations for GPT2E, and the average similarity between the entity representation and the entity mentions. We have limited the scope of the comparisons, using part-of-speech tags, to only nouns and proper nouns, as these will be the words that will be affected the most by our changes, given the dataset statistics presented in Section \ref{sec:datasets_preprocessing}.
Similarly, we calculated the average cosine similarity between the pronoun's representations of the two models as well as the differences between the two when entity representations are present.

\begin{table}[!htb]
\centering
\caption{Cosine similarity of mention representations and their entities in different scenarios}
\begin{tabular}{|c|c|c|c|}
\hline
Experiments                                                                          & \begin{tabular}[c]{@{}c@{}}GPT2E \\ without Entities\end{tabular} & \begin{tabular}[c]{@{}c@{}}GPT2E \\ with Entities\end{tabular} & GPT2    \\ \hline
\begin{tabular}[c]{@{}c@{}}Average mention similarity\\ NN,NNS,NNP,NNPS\end{tabular} & 0.7117                 & 0.7117              & 0.6971  \\ \hline
\begin{tabular}[c]{@{}c@{}}Average entity similarity\\ NN,NNS,NNP,NNPS\end{tabular}  & 0.0489                 & 0.0513              & -0.0164 \\ \hline
\begin{tabular}[c]{@{}c@{}}Average mention similarity\\ PRP,PRP\$\end{tabular}       & 0.8250                 & 0.8250              & 0.7928  \\ \hline
\begin{tabular}[c]{@{}c@{}}Average entity similarity\\ PRP,PRP\$\end{tabular}        & 0.0619                 & 0.0566              & -0.0173 \\ \hline
\end{tabular}

\label{tab:cosine_embeddings}
\end{table}

Based on the results displayed in Table \ref{tab:cosine_embeddings}, we can infer that the mentions maintain their similarity when the coreference information are used during inference, while also have a higher average similarity than the respective mentions of the model trained without coreference annotations. However, taking into account the changing similarity scores between the entity representations and the entity mentions when we use coreference information during inference, we can conclude that there is a constant change to the representations. In the case of nouns and pronouns, that change brings the representations closer while in pronouns it has the opposite effect. Individual visual representations of the embeddings for GPT2E and GPT2 and a comparative visual representation between the two are included in the appendix section.

\begin{table}[!htb]
\centering
\caption{NER performance using GPT2 and GPT2E representations as input.}
\begin{tabular}{|l|c|c|c|c|c|c|}
\hline
\multicolumn{1}{|c|}{\multirow{2}{*}{Labels}} & \multicolumn{3}{c|}{GPT2} & \multicolumn{3}{c|}{GPT2E} \\ \cline{2-7} 
\multicolumn{1}{|c|}{}                        & F1  & Prec  & Recall & F1  & Precision  & Recall  \\ \hline
PERSON & 48\% & 95.5\% & 32.5 \%& 51.5\% & 94\% & 35.5\% \\ \hline 
PRODUCT & 8\% &	33\% & 4.5 \% & 23.5\% & 90\% & 13.5\% \\ \hline 
EVENT & 23\% &	83.5\% & 13.5\% & 15\% & 75\% & 8.5\% \\ \hline 
CARDINAL & 28\% & 81.5\% &	17.5\% & 34\% & 75\% & 23\% \\ \hline
NORP & 44.5\% &	72.5\% & 36\% & 48\% & 79\% & 39.5\% \\ \hline \bottomrule 
Overall & 54\% & 87\% & 39\% & 57\% & 88\% & 42\% \\ \hline
\end{tabular}

\label{tab:ner_performance}
\end{table}

The NER model, trained using word representations from GPT2E, achieved a mean average 3\% F1 increase than the one trained with GPT2 word representations. We highlight four named entities in Table \ref{tab:ner_performance}, which showed the biggest differences between the two trained models. Specifically, we observe that the named entities of PERSON and PRODUCT, which would be directly affected by the anaphoric information in the training process, showed the greatest increase and contributed the most to the performance boost. Subsequently, EVENT entities were more commonly mislabelled while using GPT2E representations. This behaviour is credited to the use of LOCATION terms to describe events (e.g. ``the Guangzhou Fair'') and to generic event terms that refer to different entities based on their context (e.g. ``new year'' can refer to a different year) which the baseline model was unable to handle correctly when the word representations were affected by entity information. 
\section{Discussion and Conclusions}

In this paper we demonstrated a novel architecture to use coreference information in transformer-based neural language models in order to create richer representations and its effects on downstream tasks. We introduced an extension over the Transformer blocks of GPT2, labeled Entity-Transformer, that integrates coreference information to each entity mention. To that end, we also created an entity handling mechanism to create and update entity representations. Furthermore, as our proposed architecture extends over the basic Transformer block, it can be easily adapted to other Transformer-based language models, such as BERT, and also enables further research for Transformer-based language models with explicit entity decisions which have far outperformed their RNN counterparts.

In our experiments we showcased that in terms of Language modelling, both GPT2E and GPT2, when trained on the same data, have indistinguishable performance in terms of Perplexity and GPT2E has a small computational cost that translates into a slightly longer training time. However, the difference in the similarity between entity mention representations suggests that fewer iterations and mentions of each word are required to achieve the results, assuming a large enough number of mentions. This is due to the extended contextual information present at each mention occurrence, in the form of entity representations, used when training the model. What is more, the differences in these representations directly translates to an increase in tasks such as Named Entity Recognition. As coreference is ever-present in natural language, with a better ability for a language model to understand and utilize the anaphoric phenomenon in text, we expect an increased performance in other tasks such as summarization and natural language inference.

In order for language models to use coreference information, there are two requirements that need to be met. First, the models need to replace the Transformer blocks with the Entity-Transformer blocks introduced and also adopt the entity handling mechanism to make use of entity information. Second, annotated coreference information are required throughout the training corpus. While the changes described for the language models are trivial, language models require an enormous amount of training data, making it impossible to manually annotate coreference information. However, the entity handling mechanism we introduced is not affected by the lack of entity information in the training and is only boosted by the existence of them. As a result, even sparse annotations of high confidence will allow for improvements in the representations. 

In the future, we plan to extend our work, using noisy annotation provided by pretrained coreference resolvers so that we can train GPT2E to the WikiText dataset \cite{merity2018analysis}, creating a comparable model with the original GPT2 and other state-of-the-art language models in a wider range of tasks. Furthermore, we aim to expand the abilities of our current approach to be able to make explicit entity decisions, similar to the previously cited work. For that purpose, attention head probing techniques, which have been found to model some anaphoric phenomena \cite{clark-etal-2019-bert}, and transfer learning through weight initialization from a pre-trained GPT2 model will be investigated as they can contribute to significant improvements while needing less annotated training data.

\section*{Acknowledgements} 
This research is co-financed by Greece and the European Union (European Social Fund- ESF) through the Operational Programme ``Human Resources Development, Education and Lifelong Learning'' in the context of the project “Strengthening Human Resources Research Potential via Doctorate Research” (MIS-5000432), implemented by the State Scholarships Foundation (IKY).

\bibliographystyle{coling}
\bibliography{crac2020}
\clearpage

\section*{Appendix A. Embeddings visualizations}
\label{sec:appendix_a}

\begin{figure}[ht]
    \centering
    \begin{subfigure}[t]{0.45\textwidth}
        \centering
        \includegraphics[width=\linewidth]{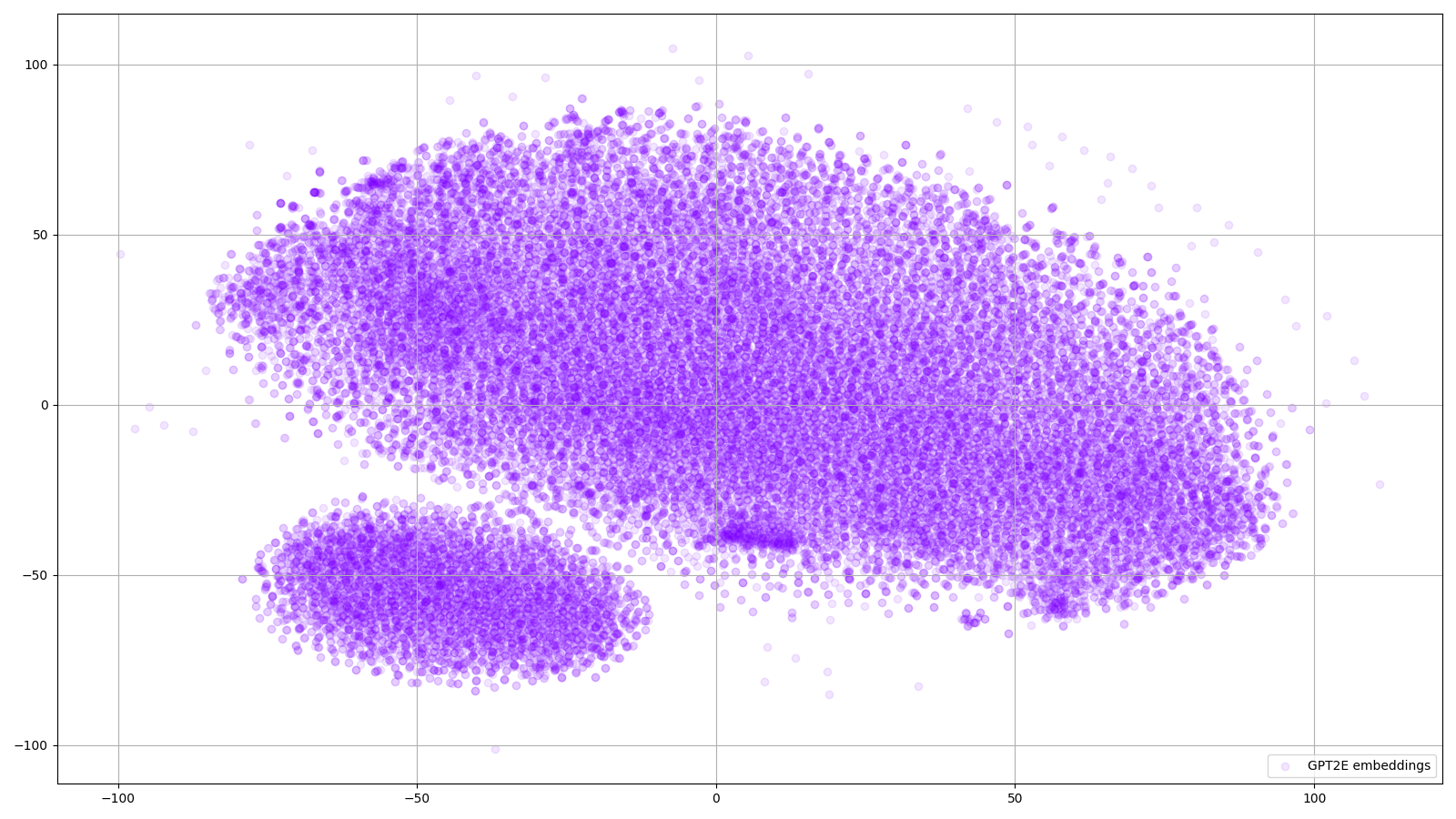} 
        \caption{GPT2E embeddings.} \label{fig:gpt2e_emb}
    \end{subfigure}
    \hfill
    \begin{subfigure}[t]{0.45\textwidth}
        \centering
        \includegraphics[width=\linewidth]{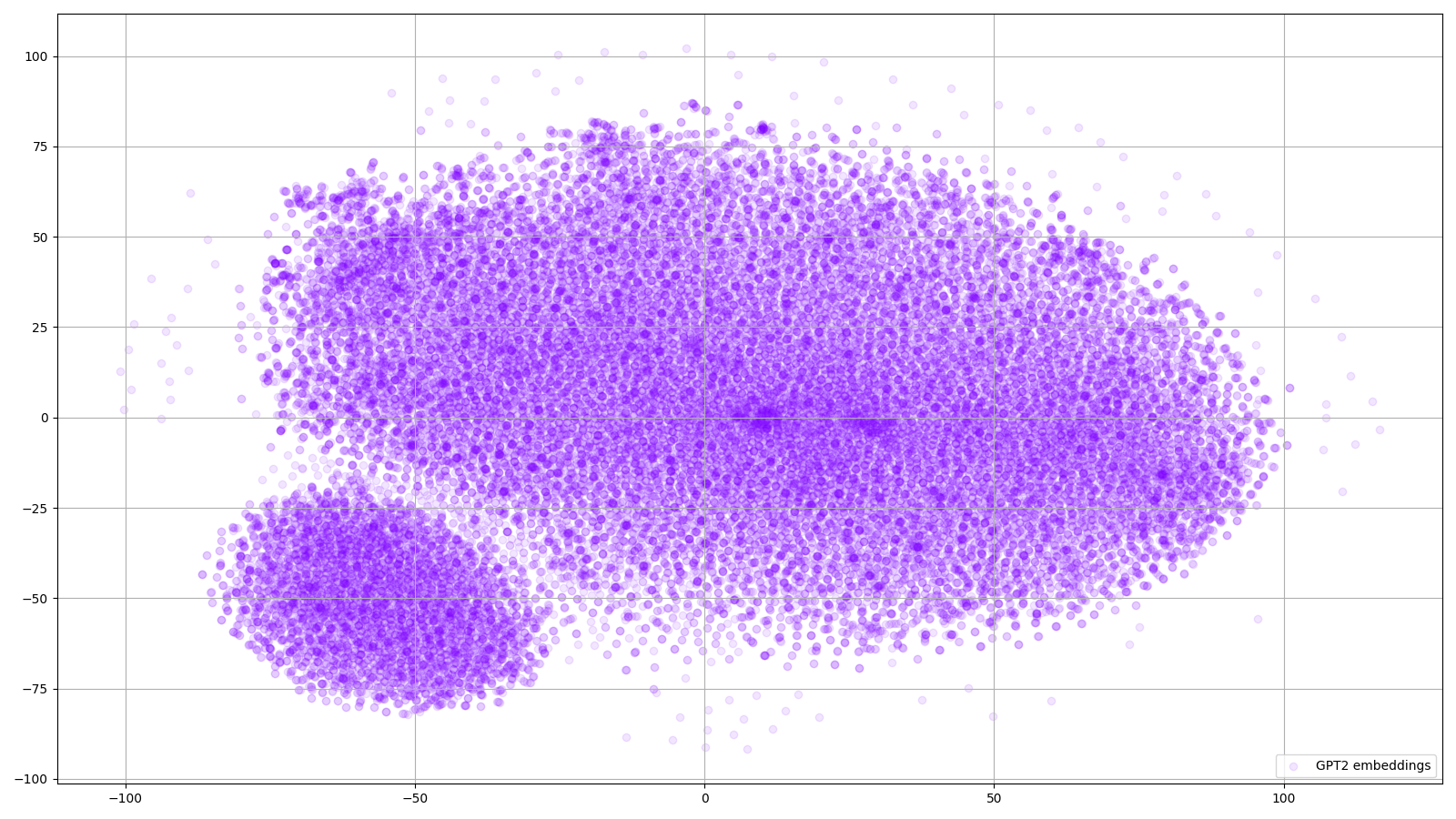} 
        \caption{GPT2 embeddings.} \label{fig:gpt2_emb}
    \end{subfigure}

    \vspace{1cm}
    \begin{subfigure}[t]{\textwidth}
    \centering
        \includegraphics[width=\linewidth]{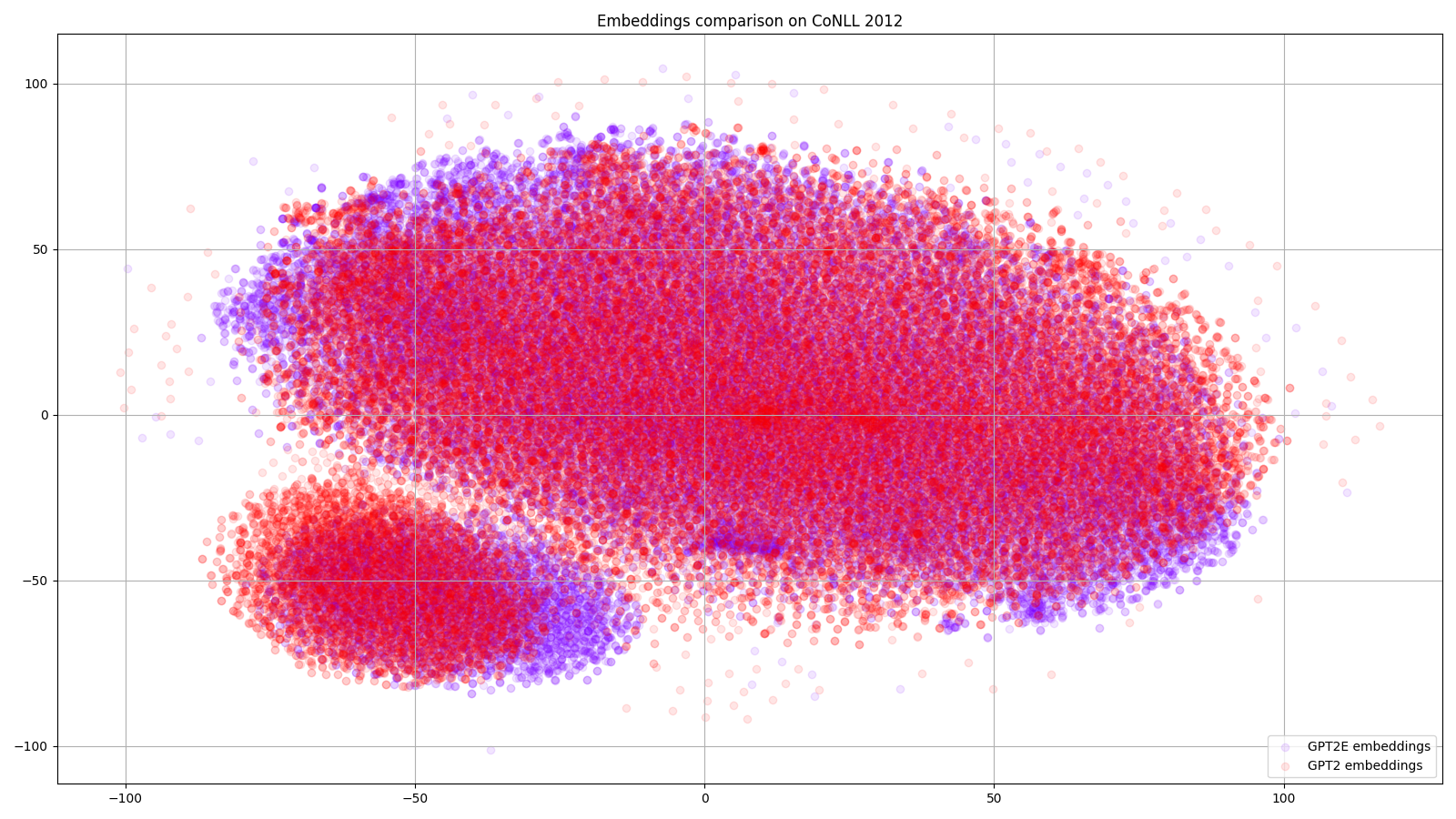} 
        \caption{Embeddings comparison between GPT2E and GPT2.} \label{fig:embs_compare}
    \end{subfigure}
    \caption{Visualization of the word representations of (a) GPT2E and (b) GPT2E and (c) comparison between the two, trained on the CoNLL2012 dataset, using t-SNE.}
\end{figure}

\end{document}